# Unsupervised Broadcast News Summarization; a comparative study on Maximal Marginal Relevance (MMR) and Latent Semantic Analysis (LSA)

Majid Ramezani
*Computerized Intelligence Systems Laboratory*
*Department of Computer Engineering*
*University of Tabriz*
Tabriz, Iran
m_ramezani@tabrizu.ac.ir

Mohammad-Salar Shahryari
*Department of Computer Engineering*
*University of Tabriz*
Tabriz, Iran
s.shahryari@tabrizu.ac.ir

Amir-Reza Feizi-Derakhshi
*Computerized Intelligence Systems Laboratory*
*Department of Computer Engineering*
*University of Tabriz*
Tabriz, Iran
derakhshi97@ms.tabrizu.ac.ir

Mohammad-Reza Feizi-Derakhshi
*Computerized Intelligence Systems Laboratory*
*Department of Computer Engineering*
*University of Tabriz*
Tabriz, Iran
mfeizi@tabrizu.ac.ir

*Abstract*— The methods of automatic speech summarization are classified into two groups: supervised and unsupervised methods. Supervised methods are based on a set of features, while unsupervised methods perform summarization based on a set of rules. Latent Semantic Analysis (LSA) and Maximal Marginal Relevance (MMR) are considered the most important and well-known unsupervised methods in automatic speech summarization. This study set out to investigate the performance of two aforementioned unsupervised methods in transcriptions of Persian broadcast news summarization. The results show that in generic summarization, LSA outperforms MMR, and in query-based summarization, MMR outperforms LSA in broadcast news summarization.

## I. Introduction

Undoubtedly, the role of information cannot be ignored in today's human life. Concerning the fact that the available information is more than what is required, and the volume of information increases exponentially, there should be some tools by which the essential information can be acquired in various information resources. In this regard, automatic summarization systems aim to extract the most essential and available information from one or more resources. They are considered a competent substitution for perusing the whole document. In addition, they cause to save a lot of time.

Research on automatic summarization began in 1950 in the field of automatic summarization of text. It has been considerably taken into account in recent years ([1]-[11]). Successes in automatic text summarization encouraged the researchers toward automatic speech summarization which refers to extracting the most critical information available in speech documents and removing redundant and irrelevant information [12]. Therefore, gaining access to the required information and searching information among great speech documents such as broadcast news, lectures, presentations, multi-party meeting recordings, spontaneous dialogues, voicemail, telephone conversations etc. have been facilitated [13]-[17].

Generally, speech summarization methods are classified into two groups. Namely, *supervised* (or *feature-based*) methods such as sentence position techniques, cue words, and acoustic/prosodic features like duration and pitch, and *unsupervised* (or *rule-based*) methods such as *Maximal Marginal Relevance (MMR)*, and *Latent Semantic Analysis (LSA)* [18].

A considerable amount of literature has been published on the speech summarization of broadcast news. In these studies, text transcripts of speech documents are available. For this purpose, both text-based and speech-based features have been utilized. Some researchers have investigated and studied speech summarization by using supervised methods. A set of linguistic features (such as word frequency, cue words, title words, and location) and non-linguistic features (such as prosodic information) have been used in [19] to summarize Japanese broadcast news. Banerjee and Rudnicky [20] have developed an extractive summarization system in multi-party human-human dialogue using text-based features. Various information, containing lexical and structural information and acoustic/prosodic information, have been used by [21] to summarize broadcast news. Other researchs, such as [22] have investigated different features like discourse, structural and lexical features, and acoustic/prosodic features to summarize English and Japanese broadcast news. By using a Support Vector Machine (SVM), authors in [23] proposed a method to summarize speech documents. The utterance importance and redundancy have also been simultaneously taken into account. In meeting summarization, Liu and Liu in [24] developed a system based on a rich set of speaker-sensitive features. Also, Wang and Cardie in [25] elaborated an abstract generation framework in a domain-independent fashion. Zhang et al. in [26] considered lecture speech summarization by using rhetorical structure.

In addition, various studies have been conducted using unsupervised methods of speech summarization. ClusterRank [27] is a graph-based system that is proposed for extractive meeting summarization; it uses an extension of the TextRank [28] algorithm (an algorithm for sentence extraction from the text). TextRank uses Google's PageRank to calculate a (relevance) score for each sentence. Authors in [29] considered a graph-based method to perform speech summarization. They combined prosodic features with an unsupervised speech summarization framework. Chen et al. [30] and Lee et al. [31] used graph-based methods in lecture summarization. Other unsupervised methods, like Language Modeling Approach (LM) and Topical Mixture Model (TMM) are used in automatic summarization to predict terms using n-grams [32]. Among unsupervised methods, LSA and MMR are considered the most well-known and widely-used methods in transcript summarization. Extractive summarization follows two objectives: maximizing the amount of covered information and minimizing redundancy [33]. MMR [34] is a greedy and query-based method that produces the final summary through sentence-to-sentence selection, considering two criteria: *similarity to the user query* and *minimum redundancy in selected sentences*. In the same manner, spoken dialogue has been summarized by [35] after the following steps: speech disfluency and removal, sentence boundary detection, identification, and linking query-answer regions, and topic segmentation. The best query has been followed by [36] in the MMR method by identifying and extracting key phrases.

In LSA [37], the implicit semantic relations of the words are identified and acquired in terms of contextual usage. LSA has been used and applied by [38], and it has been considered as a method that can identify the most important topics of the text without using a lexical thesaurus, such as WordNet, for summarization of single and multi-documents of broadcast news. It has also been used by [39] in terms of story segmentation of Chinese broadcast news. In addition, it has been used by [40] in terms of Information Retrieval (IR) in Malay broadcast news to reduce the negative effects of synonyms in information retrieval and to identify the story boundaries. Hofmann has proposed a technique called Probabilistic LSA [41]. As can be inferred, it has been derived from the LSA method. This technique involves a stronger statistical basis than LSA.

This study aims to perform unsupervised broadcast news summarization using MMR and LSA methods and evaluate their success. For this purpose, Persian broadcast news is used. In section 2, MMR and LSA are explained, and their theories are studied in detail. Then, experimental results obtained from the implementation of an automatic summarization system in broadcast news are presented in section 3, and the obtained results will be analyzed. Finally, conclusions are presented in section 4.

## II. METHODOLOGY

In this study, the summarization of broadcast news will be presented by using MMR and LSA. Generally, summarizations can be performed in both *generic* and *query-based* ways. Generic summarizations include extracting the most important information available in a document. Summaries produced using generic summarization are considered a complete substitution of original documents. In contrast, in query-based summarizations, summaries are presented based on the required information of the user determined by the query. In the following sections, summarizations using LSA or MMR methods are investigated in terms of generic and query-based methods.

### A. Maximal Marginal Relevance (MMR)

Maximal Marginal Relevance (MMR) is one of the most well-known methods in information retrieval to balance *relevance* and *diversity* [42]. This method was proposed by Carbonell and Goldstein [34] for the first time. It selects the sentences based on a weighted combination of their relevance to the input query and their similarity to the sentences that have been selected before. The idea behind this method is the iterative ranking of sentences according to the input query. In each iteration, a sentence is selected in which: it has not been selected before, it is more similar to the query, and it is more dissimilar to those previously selected sentences. According to this method, the MMR score is calculated by (1) for each $S_i$ sentence of the input document.

$$Score^{MMR}(S_i) = argmax_{S_i} \{ \lambda (Sim_1(S_i, Q)) - (1-\lambda)(Sim_2(S_i, Summ)) \} \quad (1)$$

where $Q$ is the query vector, $Summ$ is the vector of those sentences selected to appear in the final summary, and $\lambda$ is a coefficient between $0$ and $1$, which controls the amount of redundancy and similarity to the input query (relevance) in each selection. $Sim_1$ measures the similarity between the user query and current sentences, while $Sim_2$ evaluates the similarity of the current sentence with those sentences that have been selected up to now. In this method, the cosine similarity function has been used as a similarity metric for $Sim_1$ and $Sim_2$. The sentences with the highest MMR score will be iteratively selected to appear in the summary until it reaches a predefined proper size. In this regard, the $\lambda$ coefficient has great importance, and the quality of results depends on the value of this coefficient. As it can be inferred, there is a trade-off between similarity to the query (relevance) and diversity. This relation is affected by the value of $\lambda$. If $\lambda=1$, then the relevance reaches its maximum value. If $\lambda=0$, then diversity will be maximized.

Despite that, MMR basically is a query-based method, it can be used as a generic method [43]. In such a case, the document vector ($D$) substitutes the query vector ($Q$) in (1).

### B. Latent Semantic Analysis (LSA)

Latent semantic Analysis (LSA) is a method that is used to extract and represent the contextual usage of words [44]. This way, semantic similarities among the words and other sets of words are identified. LSA is a vector space method that can extract word-word, word-passage, and passage-passage relations. There are high correlations among these relations, human cognitive events, and semantic representation [45]. This method can considerably extract and identify semantic relations formed in the speaker or writer's mind during writing and speaking. Also, it can be mentioned that LSA allows us to provide semantic similarities among words for the machine through proper approximation of human judgment and to predict semantic relations between various text parts.

LSA is a full-automatic mathematical-statistical method composed of two basic steps. The first step involves representing the input text by a matrix whose rows are related to the input text's unique words. The columns are dedicated to sentences, passages, or other text units (such as paragraphs or documents). In the first step, LSA creates a *term-by-sentence* matrix ($A$), which is a matrix representation of the input text. Each column vector of the matrix ($A_i$) is an indication of a weighted term-frequency vector for the $i^{th}$ sentence. Having $t$ unique words/terms and $s$ sentences in the input text, $A$ would be a $t \times s$ matrix ($t \gg s$). Each cell of the matrix involves the frequency of the words/terms in the corresponding sentence. Of course, there are various approaches for determining matrix cells, such as the number of occurrences, the binary representation of the number of occurrences, TF- IDF (Term Frequency-Inverse Document Frequency), Log Entropy, Root Type, and Modified TF-IDF.

The second step of LSA is applying the Singular Value Decomposition (SVD) over matrix *A*. As a consequence of applying SVD, matrix *A* is decomposed into three matrices, namely, *V*, *Σ* and *U*.

$$A_{t \times s} = U_{t \times c} \ \Sigma_{c \times c} \ V^T_{c \times s} \quad (2)$$

Where *U* is a *t×c* and column-orthogonal matrix whose columns are called *left singular values*. $\Sigma = diag(\delta_1, \delta_2, ...., \delta_c)$ is a diagonal and *c×c* matrix whose main diagonal elements are *eigenvalues* of *A*. The eigenvalues have been sorted in descending order on the main diagonal. $V^T$ matrix is an orthogonal *c×s* matrix whose columns are called *right singular values*. If *rank(A)=r*, then the following relation is presented for the matrix *Σ*:

$\sigma_1 \geq \sigma_2 \geq \sigma_3 \geq .... \geq \sigma_r \geq \sigma_{r+1} = ...... = \sigma_s = 0$

Matrices obtained from applying the SVD operator to the input matrix can be interpreted as follows [46]: each column in *U*, which is a *t×c* matrix, indicates a unique concept or topic of the input text. In this matrix, the rows are the words or terms available in the original space. Each cell indicates the weight of the corresponding term in the corresponding concept or topic. The diagonal matrix *Σ* with *c×c* dimensions involves the significance of each concept or topic that appeared in the original text. Since the cells are sorted in descending order, the concepts with less importance can be ignored by reduction. Moreover, $V^T$ is a new representation of sentences in the input text. The difference is that, in this representation, sentences are not represented using words and terms in the original input text, but they are represented using the terms of the topics given in *U*. Keeping in mind that although that LSA is a generic method, it can also be used as a query-based method. They are described in detail below.

*1) Generic LSA*

Automatic summarization using LSA was proposed by Gong and Liu for the first time [38]. It was also improved by Steinberger and Jezek [47] and Murray et al. [48]. As described below, all of them perform generic summarization and involve the two main steps of the LSA along with a *sentence selection* step. Sentence selection is the central focus of all LSA-based methods in the current study.

*a) Gong and Liu*

The study by Gong and Liu [55] is considered the most crucial in LSA-based automatic summarization. According to their suggestion, the final summary is acquired after the creation of *t×s* matrix and decomposing it by applying the SVD. Then the summary will be acquired using a sentence selection step which primarily relies on the $V^T$ matrix. According to their algorithm to select sentences, the first concept (that is, the most important concept, which is placed in the first row of $V^T$) is selected first. Then, the sentence most relevant to this concept (the sentence with the highest value in the corresponding row) will be presented in the final summary. It continues for other concepts so that a predefined number of sentences are extracted. Fig. 1, demonstrates an example of $V^T$ matrix. According to this algorithm, the *Con. 0* concept is first selected, and then $S_1$ (the cell with the highest value in *Con.* 0) is selected for the final summary.

|        | $S_0$ | $S_1$ | $S_2$ | $S_3$ | $S_4$ |
|--------|-------|-------|-------|-------|-------|
| Con. 0 | 0.557 | 0.691 | 0.241 | 0.110 | 0.432 |
| Con. 1 | 0.345 | 0.674 | 0.742 | 0.212 | 0.567 |
| Con. 2 | 0.732 | 0.232 | 0.435 | 0.157 | 0.246 |
| Con. 3 | 0.628 | 0.836 | 0.783 | 0.265 | 0.343 |

Fig. 1. An example of $V^T$ matrix (Gong & Liu). From each row of $V^T$, the sentence with the highest score is selected until a predefined number of sentences are selected.

|       | Con. 0 | Con. 1 | Con. 2 | Con. 3 | Length |
|-------|--------|--------|--------|--------|--------|
| $S_0$ | 0.846  | 0.334  | 0.231  | 0.210  | 0.432  |
| $S_1$ | 0.455  | 0.235  | 0.432  | 0.342  | 0.543  |
| $S_2$ | 0.562  | 0.632  | 0.735  | 0.857  | 0.723  |
| $S_3$ | 0.378  | 0.186  | 0.248  | 0.545  | 0.235  |

Fig. 2. An example of *V* matrix (Steinberger & Jezek). For each row of *V*, the length of the sentences using *n* concepts is calculated. *n* is given as an input parameter. The values in *Σ* are used as importance while achieving the lengths.

|        | $S_0$ | $S_1$ | $S_2$ | $S_3$ | $S_4$ |
|--------|-------|-------|-------|-------|-------|
| Con. 0 | 0.557 | 0.691 | 0.241 | 0.110 | 0.432 |
| Con. 1 | 0.345 | 0.674 | 0.742 | 0.212 | 0.567 |
| Con. 2 | 0.732 | 0.232 | 0.435 | 0.157 | 0.246 |
| Con. 3 | 0.628 | 0.836 | 0.783 | 0.265 | 0.343 |

Fig. 3. An example of $V^T$ matrix (Murray et al.). From each row of $V^T$, one or more sentences with higher scores are selected. The number of sentences to be selected is determined by using *Σ*.

The algorithm suffers from some defects. Firstly, the size of the reduced dimension should be equal to the final summary's length. Therefore, selecting some sentences with less significance is probable. Secondly, some sentences are relevant to the selected concept, but the related cell does not have the highest value in the corresponding row. According to this algorithm, such sentences cannot be presented in summary (while such sentences result in increasing the relevance in the final summary). Thirdly, unlike the LSA, the significance of all concepts is considered equally, while some of them don't have great importance in the original document.

*b) Steinberger and Jezek*

Steinberger and Jezek [47] proposed an algorithm to improve the defects of the method suggested by Gong and Liu. In this algorithm, after performing two basic steps of LSA, *Σ* and *V* matrices have been used to select the sentences. In this algorithm, the sentence selection begins with calculating the length of sentence vectors placed in the rows of the *V*. To calculate the length of a sentence vector, the concepts whose index is less than or equal to the number of dimensions in the new space are used. The size of new space dimensions is considered as input in this algorithm. *Σ* is a diagonal matrix containing eigenvalues of matrix *A*, which are sorted in descending order in the main diagonal. This matrix conceptually involves the amount of importance of the concepts proposed in the original document that has been transferred to the reduced dimension space. Therefore, by considering the values in this matrix as a multiplication parameter, these concepts that are more relevant to the document will have great importance. If the dimension of the new space is equal to *n*, then the length of the $i^{th}$ sentence is calculated according to (3).

$$Length \ i = \sqrt{\sum_{j=1}^{n} V_{ij} \times \Sigma_{jj}} \quad (3)$$

After calculating the length of all sentences, the longest sentences are selected for the summary. As it can be inferred, the selection of more than one sentence from each concept is possible in this algorithm. Sentences that are more relevant to significant concepts are selected as well. It's evident that in this algorithm, the significance of various concepts extracted by SVD is not considered equal. Fig. 2, shows a sample $V$ matrix. If the dimension of the new space is equal to *3*, then the length of sentences is calculated by considering the first *3* concepts, and finally, $S_2$ is selected as the first sentence.

*c) Murray et al.*

Like the previous algorithm, this algorithm was proposed by Murray et al. [48] to solve the problems of the algorithm proposed by Gong and Liu. In this algorithm, two steps of the LSA method are performed before the sentence selection. This algorithm is similar to the algorithm proposed by Gong and Liu, but the main difference is that here, *n* number of the best sentences are selected instead of selecting the best sentence from each concept. The value of *n* is calculated by singular values in the matrix $\Sigma$. In this algorithm, $\Sigma$ and $V^T$ are used in sentence selection. The number of sentences selected from each concept is determined by calculating the percentage of corresponding singular value in $\Sigma$ over the sum of all singular values in $\Sigma$. Therefore, selecting more than one sentence from each concept is possible. In addition, the significance of various extracted concepts is not considered equal. In addition, dimension reduction is not dependent on the length of the final summary. Fig. 3, shows a sample $V^T$ matrix. Suppose that two sentences must be selected from *Con. 0*, and one sentence must be selected from the *Con. 1*. According to this algorithm, $S_0$ and $S_1$ are selected from *Con. 0*, and $S_2$ is selected from *Con. 1*.

*2) Query-based LSA*

In the query-based approach, the obtained results are affected by the user query. Generally, the method of literally matching the terms in the user's query and terms in documents can apply user query to the information retrieval. But it should be noted that literal or lexical matching and correspondence involve incorrect results [37]. Usually, there are different methods for stating a concept. Concerning this issue, such documents can not be identified and considered relevant. In addition, most of the words are polysemous. Therefore, it's possible to compare the words of the user's query with the words of irrelevant documents. In the query-based LSA to solve the problems of lexical matching, documents as well as the user's query, are projected to the new space. In this new space, there may be a high cosine similarity between a query and a document without any common word. Because according to LSA, the words of query and sentences may be semantically similar. It can be stated that the LSA method represents documents and queries in reduced dimensional space by using the SVD operator instead of representing them based on vectors with independent words. Since the number of factors or dimensions in the new space is less than when the words are separately represented, the words are not independent. For instance, if two words are presented in a similar context, they will be represented by a single vector in reduced dimensional space. In this regard, even sentences (documents) that do not share words with queries can be retrieved. Like the generic approach, in the query-based LSA, at first, the processed text is represented by matrix *A*. Then, it is projected to a latent semantic space by applying the SVD (refer to (2)).

Finally, the user query *q* should be transferred to the new space using (4). It also provides a logical and mathematical basis for evaluating the similarity of query vectors ($\hat{q}$) and sentence vectors in new semantic space.

$$\hat{q}= q^T\ U_{t\times r} \sum\nolimits^{-1}_{r\times r} \qquad (4)$$

where *q* is the $t\times 1$ vector of the user's query. Then, multiplying $q^T_{1\times t}\ U_{t\times r}$, the query vector is transferred from the original space to the new space. Finally, multiplying it by $\sum^{-1}_{r\times r}$, suitable weights are assigned to each dimension. In this regard, the required basis for comparing query vectors with sentences' vectors is provided in new space. For this purpose, *n* number of sentences in the input text whose cosine similarity is higher than the required threshold is selected for the summary.

*3) Probabilistic LSA (PLSA)*

Probabilistic Latent Semantic Analysis (PLSA) [41] is a method obtained from the LSA technique by emphasizing statistical aspects. In other words, this technique has a stronger statistical basis by emphasizing the Maximum Likelihood Principle. PLSA uses Hofmann's Aspect Model. Based on this method, a set of unobserved variables, $z\epsilon Z=\{z_1, z_2,..., z_k\}$ that appeared topics in documents are related to two sets of observed variables including documents $d\epsilon D=\{d_1, d_2, ..., d_m\}$ that here are considered as sentences, and the words of these documents $w\epsilon W=\{w_1, w_2, ... w_n\}$. In this method the probability of each of the co-occurrences is computed using (5) by observing the co-occurrence of the words and documents $(w, d)$:

$$P(d, w)= P(d)\ P(w|d)\ ,\ P(w|d)=\sum\nolimits_{(z\epsilon Z)} P(w|z)P(z|d)$$
$$P(d, w)= P(d) \sum\nolimits_{z\in Z} P(w|z)P(z|d) \qquad (5)$$

where $P(d)$ is the probability of selecting document *d*, $P(w|z)$ is the probability of selecting the word *w* from topic *z*, $P(z|d)$ stands for the probability of selecting topic *z* in document *d*, and $P(d, w)$ shows the likelihood of each document and word pair. In the next step, the values of $P(d)$, $P(w|z)$ and $P(z|d)$ are optimized by using the maximization of log-likelihood based on the Maximum Likelihood Principle, and the maximum value is determined. For this purpose, the expectation-maximization algorithm can be used. This algorithm is an iterative method in statistics, and it is used to determine the Maximum Likelihood when the model depends on a set of unobserved latent variables [41]. This algorithm involves two steps: the expectation step (*E-step*), in which the previous probabilities related to latent variables are computed, and the Maximization step (*M-step*), in which parameters are updated. Hofmann has proposed the (6) for E-step:

$$P(z|d, w)=\frac{P(z)P(d|z)P(w|z)}{\sum_{\acute{z}\in Z} P(\acute{z})P(d|\acute{z})P(w|\acute{z})} \qquad (6)$$

In addition, he has suggested the (7), (8) and (9) for M-step:

$$P(w|z) = \frac{\sum_d n(d,w)\ P(z|d,w)}{\sum_{d,w'} n(d,w')P(z|d,w')} \qquad (7)$$

$$P(d|z) = \frac{\sum_w\ n(d,w)P(z|d,\ w)}{\sum_{d',w}\ n(d',w)P(z|d',w)} \qquad (8)$$

$$P(z) = \frac{\sum_{d,w} n(d,w)P(z|d.w)}{\sum_{d,w} n(d,w)} \qquad (9)$$

where, $n(d, w)$ is the frequency of *w* in *d*.

E-step includes computing the probability of that word $w$ present in document $d$ can be explained by the factor $z$ (topic). Then, the converging point can be found by alternating the E-step and M-step. This point has the role of local maximum in log-likelihood.

The PLSA-based summarization algorithm involves the following steps:
i.  Each document is represented in terms of the term-frequency matrix.
ii. The values of $P(z)$, $P(d|z)$ and $P(w|z)$ are calculated as declared in (7), (8), and (9) until the convergence criterion for EM-algorithm is obtained. $P(d|z)$ shows the importance of document $d$ (sentence) in the given topic represented by $z$. $P(z)$ represents the importance of topic $z$ in document $d$.
iii. R score is computed by using the (10) for each sentence. In this case, all sentences are scored according to their topic inclusion.

$$R=\sum_z P(d|z)P(z) = P(d) \qquad (10)$$

iv. Finally, the sentences with the highest $R$ score are selected to form the summary.

## III. EXPERIMENTS AND RESULTS

### A. Dataset

The corpus used in this study results from the human transcription of Persian broadcast news for more than 15 hours, containing 58 news documents. Four native transcribers have collected it from the news of three radio channels and four TV channels over 45 days. The corpus involves more than 115,000 words and 7,000 sentences. The average number of sentences per news document is 122.5, and the average length of sentences in the corpus is 16.5. To produce the golden summaries as the references to evaluate summarization systems, four human summarizers summarized all the documents, for both generic and query-based summarization (for pre-specified queries). The average number of sentences per golden summary is equal to 36.7.

### B. Results and Discussions

To evaluate the proposed summarization systems, we used *Precision*, *Recall* and *F1-score*, along with the *ROUGE-1* with a compression rate equal to 30%.

$$Precision = \frac{TP}{Tp+FP} \qquad (11)$$

$$Recall = \frac{TP}{TP+FN} \qquad (12)$$

$$F1\text{-}score = \frac{2 \times Precision \times Recall}{Precision+Recall} \qquad (13)$$

$$ROUGE\text{-}n = \frac{\sum_{S\in\{ReferenceSummaries\}}\sum_{gram_n\in S} Count_{match}(gram_n)}{\sum_{S\in\{ReferenceSummaries\}}\sum_{gram_n\in S} Count(gram_n)} \qquad (14)$$

where, in (11) and (12) *TP*, *FP*, and *FN* respectively, indicate the number of true positives, false positives, and false negatives. In (14), *n* stands for the length of the n-gram (in this study $n=1$), $Count_{match}(gram_n)$ is the maximum number of n-grams co-occurring in a candidate summary, and $Count(gram_n)$ is the number of n-grams in the reference summary.

#### 1) Summarization Using MMR

As mentioned before, in MMR documents are summarized to provide a balance between relevance and redundancy. It is followed by setting $\lambda$ (see (1)). The highest value of $\lambda$ will cause more emphasis on relevance. And the lowest value of $\lambda$, will cause more diversity. In this study, we considered the results obtained from this method by assigning different values of {0, 0.1, 0.2, ..., 1.0} to $\lambda$. Table I demonstrates evaluation results obtained by using query-based MMR for four evaluation metrics. The applied queries are predefined for each news document in the dataset.

TABLE I. EVALUATION RESULTS FOR QUERY-BASED MMR

| λ | Precision (%) | Recall (%) | F-score (%) | ROUGE-1 |
|---|---|---|---|---|
| 0.0 | 33.34 | 21.87 | 26.22 | 0.2437 |
| 0.1 | 35.72 | 23.54 | 28.22 | 0.2494 |
| 0.2 | 35.72 | 23.54 | 28.22 | 0.2267 |
| 0.3 | 65.87 | 46.25 | 54.17 | 0.4310 |
| 0.4 | 71.43 | 49.28 | 58.09 | 0.4487 |
| 0.5 | 61.90 | 42.05 | 49.88 | 0.3982 |
| 0.6 | 76.19 | 49.26 | 59.56 | 0.4847 |
| 0.7 | 72.22 | 48.79 | 57.97 | 0.4839 |
| 0.8 | 69.84 | 46.94 | 55.88 | 0.4681 |
| 0.9 | 69.84 | 46.94 | 55.88 | 0.4681 |
| 1.0 | **83.34** | **54.66** | **65.69** | **0.5379** |

TABLE II. EVALUATION RESULTS FOR GENERIC MMR

| λ | Precision (%) | Recall (%) | F-score (%) | ROUGE-1 |
|---|---|---|---|---|
| 0.0 | 22.22 | 20.04 | 21.06 | 0.2082 |
| 0.1 | 24.61 | 22.26 | 23.36 | 0.2249 |
| 0.2 | 19.84 | 18.10 | 18.92 | 0.1678 |
| 0.3 | 37.30 | 33.13 | 35.08 | 0.3345 |
| 0.4 | 38.10 | 33.45 | 35.61 | 0.3227 |
| 0.5 | 33.32 | 29.56 | 31.32 | 0.2921 |
| 0.6 | 26.19 | 23.18 | 24.58 | 0.2275 |
| 0.7 | **40.48** | **35.68** | **37.91** | **0.3807** |
| 0.8 | **40.48** | **35.68** | **37.91** | **0.3807** |
| 0.9 | **40.48** | **35.68** | **37.91** | **0.3957** |
| 1.0 | **40.48** | **35.68** | **37.91** | **0.3957** |

As it can be seen, by increasing $\lambda$, all of the evaluation metrics were improved so that the best results were obtained in $\lambda=1$. Therefore, it can be inferred that the best results are obtained in broadcast news summarization by using query-based MMR when just the similarity of selected sentences with the query is considered ($\lambda=1$).

Table II demonstrates the results obtained from the evaluation of summaries produced by using generic MMR for four evaluation metrics. As can be seen, like generic MMR, by increasing $\lambda$, the performance of the system improves so that the best results are obtained from the maximum value of $\lambda$ (0.7, 0.8, 0.9, and 1.0). Comparing the query-based MMR and generic MMR reveals that query-based MMR generally outperforms generic MMR.

#### 2) Summarization Using LSA

Table III shows the evaluation results of broadcast news using generic and query-based LSA. Among generic approaches of LSA, the algorithm proposed by Murray et al. has achieved the best results. As can be seen, generally generic LSA outperforms the query-based LSA in all evaluation metrics. Table IV depicts the evaluation results for PLSA-based summarization. Summarization was done for various values of $n= 2,3,4,5,6$. Since in $n=1$, only the sentences of one topic are selected for the final summary, the value of $n=1$ has not been taken into account. As can be observed, increasing the value of n until $n=3$ leads to achieving the best results.

#### 3) MMR vs LSA: Which Is the Best Approach in Broadcast News Summarization?

Table V compares generic LSA and MMR, and Table VI compares query-based LSA and MMR. As it is observed, between generic MMR and query-based MMR, better results are obtained in query-based MMR in terms of broadcast news summarization. In addition, between generic LSA and query-based LSA, better results are obtained in generic LSA. On the other hand, regarding the fact that MMR is inherently a query-based method and LSA is inherently a generic method, the best results are respectively obtained in query-based and generic approaches.

TABLE III. EVALUATION RESULTS FOR GENERIC LSA AND QUERY-BASED LSA

| | Method | Precision (%) | Recall (%) | F-score (%) | ROUGE-1 |
|---|---|---|---|---|---|
| Generic | Gong & Liu | 55.96 | 49.39 | 52.46 | 0.5737 |
| | Steinberger & Jezek (n=1) | 57.14 | 50.68 | 53.70 | 0.5339 |
| | Murray et al. | **60.32** | **53.35** | **56.61** | **0.5930** |
| | **Query-Based** | 54.76 | 42.59 | 47.08 | 0.4854 |

TABLE IV. EVALUATION RESULTS FOR PLSA

| n | Precision (%) | Recall (%) | F-score (%) | ROUGE-1 |
|---|---|---|---|---|
| 2 | **58.31** | 28.96 | 38.39 | 0.3167 |
| 3 | 43.32 | **38.08** | **40.48** | **0.4012** |
| 4 | 53.33 | 23.97 | 32.57 | 0.2507 |
| 5 | 50.83 | 21.47 | 29.56 | 0.2258 |
| 6 | 48.32 | 19.48 | 38.68 | 0.2035 |

TABLE V. COMPARING GENERIC LSA WITH GENERIC MMR

| Method | Precision (%) | Recall (%) | F-score (%) | ROUGE-1 |
|---|---|---|---|---|
| LSA (Murray et al.) | **60.32** | **53.35** | **56.61** | **0.5930** |
| MMR ($\lambda$=1.0) | 40.48 | 35.68 | 37.91 | 0.3957 |

TABLE VI. COMPARING QUERY-BASED LSA WITH QUERY-BASED MMR

| Method | Precision (%) | Recall (%) | F-score (%) | ROUGE-1 |
|---|---|---|---|---|
| LSA | 54.76 | 42.59 | 47.08 | 0.4854 |
| MMR ($\lambda$=1.0) | **83.34** | **54.66** | **65.69** | **0.5379** |

## IV. CONCLUSION

In this study, we investigated unsupervised broadcast news summarization. MMR and LSA are the most well-known unsupervised methods in automatic speech summarization that are widely used in summarization. We suggested several systems to perform generic and query-based automatic broadcast news summarization that was based on Maximal Marginal Relevance (MMR) and Latent Semantic Analysis (LSA). The results demonstrated that MMR is preferred to LSA for query-based summarization, and LSA is preferred to MMR for generic summarization.